\documentclass[10pt,twocolumn,letterpaper]{article}
\usepackage[T1]{fontenc}
\usepackage[utf8]{inputenc}
\usepackage{authblk}
\usepackage{cvpr}              

\usepackage{graphicx}
\usepackage{amsmath}
\usepackage{amssymb}
\usepackage{booktabs}
\usepackage{makecell}
\usepackage{multirow}
\usepackage{indentfirst}
\usepackage{hyperref}
\hypersetup{hidelinks,
  colorlinks=true,
  citecolor = green,
  urlcolor = magenta,
  pdfstartview=Fit,
  breaklinks=true}
\usepackage[capitalize]{cleveref}
\crefname{section}{Sec.}{Secs.}
\Crefname{section}{Section}{Sections}
\Crefname{table}{Table}{Tables}
\crefname{table}{Tab.}{Tabs.}

\def\cvprPaperID{2} 
\def\confName{CVPR}
\def\confYear{2023}

\begin{document}

\title{MM-BSN: Self-Supervised Image Denoising for Real-World with Multi-Mask based on Blind-Spot Network}

\author[1\thanks{Corresponding author}]{Dan Zhang}
\author[1]{Fangfang Zhou}
\author[1]{Yuwen Jiang}
\author[2]{Zhengming Fu}
\affil[1]{Senslab Technology, Shanghai, China}
\affil[2]{NeuroSens Technology, Austin, U.S.}
\renewcommand\Authands{ and }
\affil[Emails:]{
	{\tt\small {zhang.dan, zhou.fangfang, jiang.yuwen@senslab.com,}}
	{\tt\small zenith.fu@neurosens.ai}
}


\maketitle
\begin{abstract}
	Recent advances in deep learning have been pushing image denoising techniques to a new level. In self-supervised image denoising, blind-spot network (BSN) is one of the most common methods. However, most of the existing BSN algorithms use a dot-based central mask, which is recognized as inefficient for images with large-scale spatially correlated noise. In this paper, we give the definition of large-noise and propose a multi-mask strategy using multiple convolutional kernels masked in different shapes to further break the noise spatial correlation. Furthermore, we propose a novel self-supervised image denoising method that combines the multi-mask strategy with BSN (MM-BSN). We show that different masks can cause significant performance differences, and the proposed MM-BSN can efficiently fuse the features extracted by multi-masked layers, while recovering the texture structures destroyed by multi-masking and information transmission. Our MM-BSN can be used to address the problem of large-noise denoising, which cannot be efficiently handled by other BSN methods. Extensive experiments on public real-world datasets demonstrate that the proposed MM-BSN achieves state-of-the-art performance among self-supervised and even unpaired image denoising methods for sRGB images denoising, without any labelling effort or prior knowledge. Code can be found in \href{https://github.com/dannie125/MM-BSN}{https://github.com/dannie125/MM-BSN}.
\end{abstract}
\label{sec:intro}
\begin{figure}[htbp]
	\setlength{\abovecaptionskip}{0cm}
	\setlength{\belowcaptionskip}{-1pt}
	\centering
	\begin{subfigure}{0.495\linewidth}
		\includegraphics[width=1\linewidth]{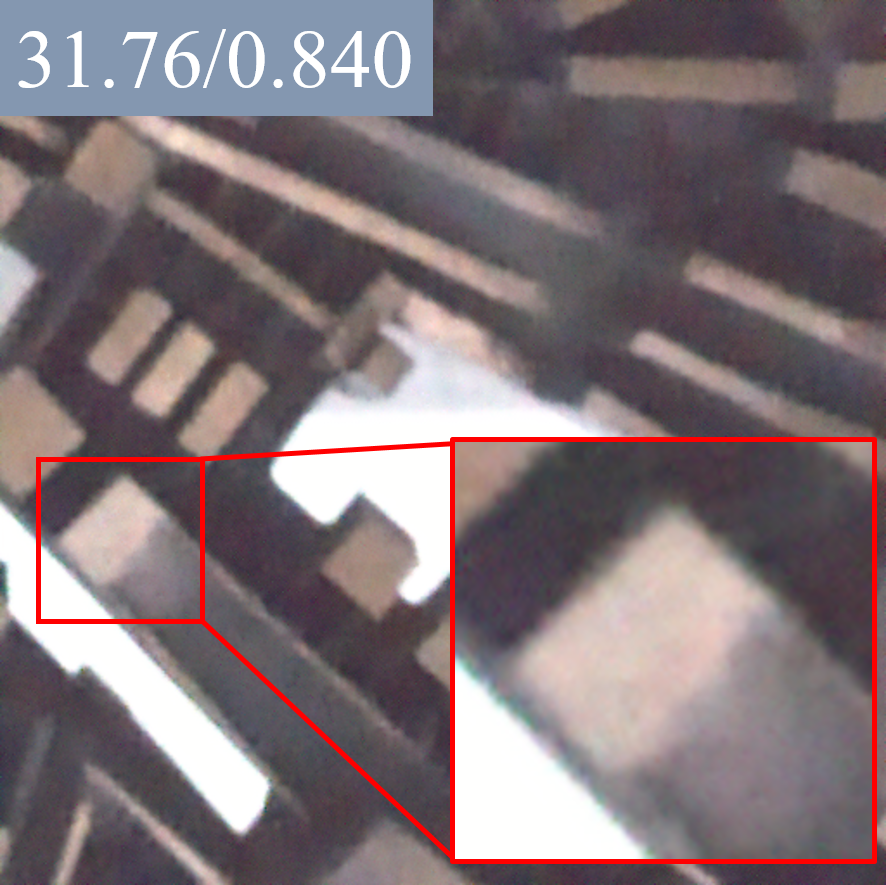}
		\caption{DnCNN \cite{zhang2017beyond}\\(Supervised)}
		\label{fig:1-a}
	\end{subfigure}
	\hfill
	\begin{subfigure}{0.495\linewidth}
		\includegraphics[width=1\linewidth]{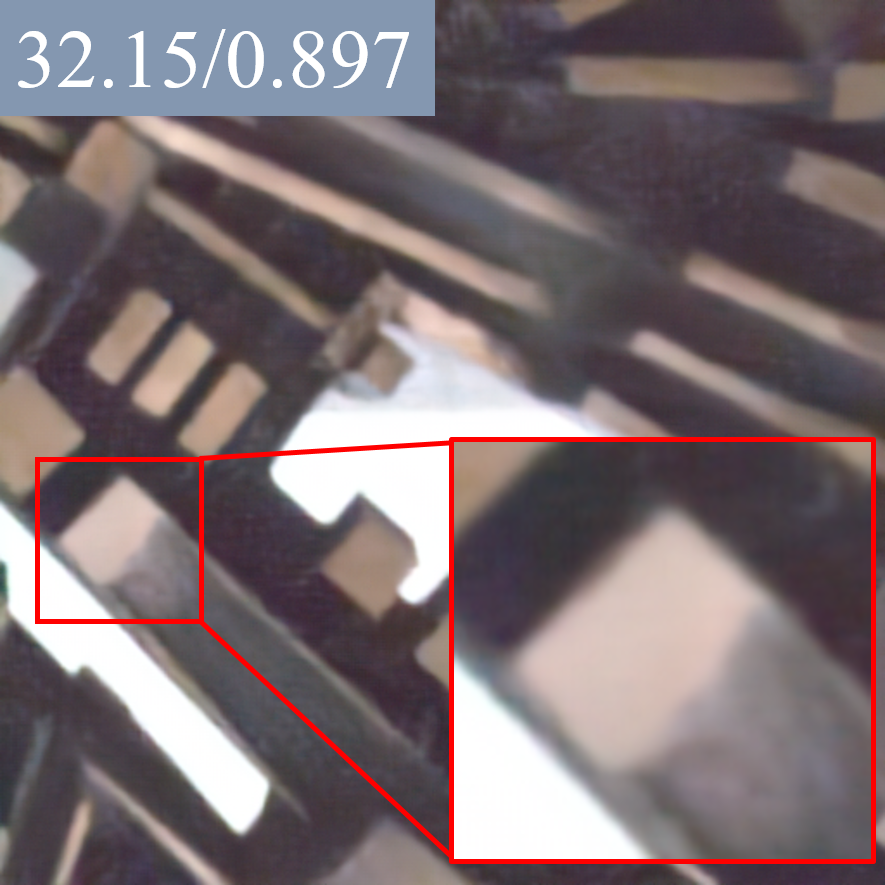}
		\caption{C2N \cite{jang2021c2n}+DIDN \cite{yu2019deep}\\(Unpaired)}
		\label{fig:1-b}
	\end{subfigure}
	\begin{subfigure}{0.495\linewidth}
		\includegraphics[width=1\linewidth]{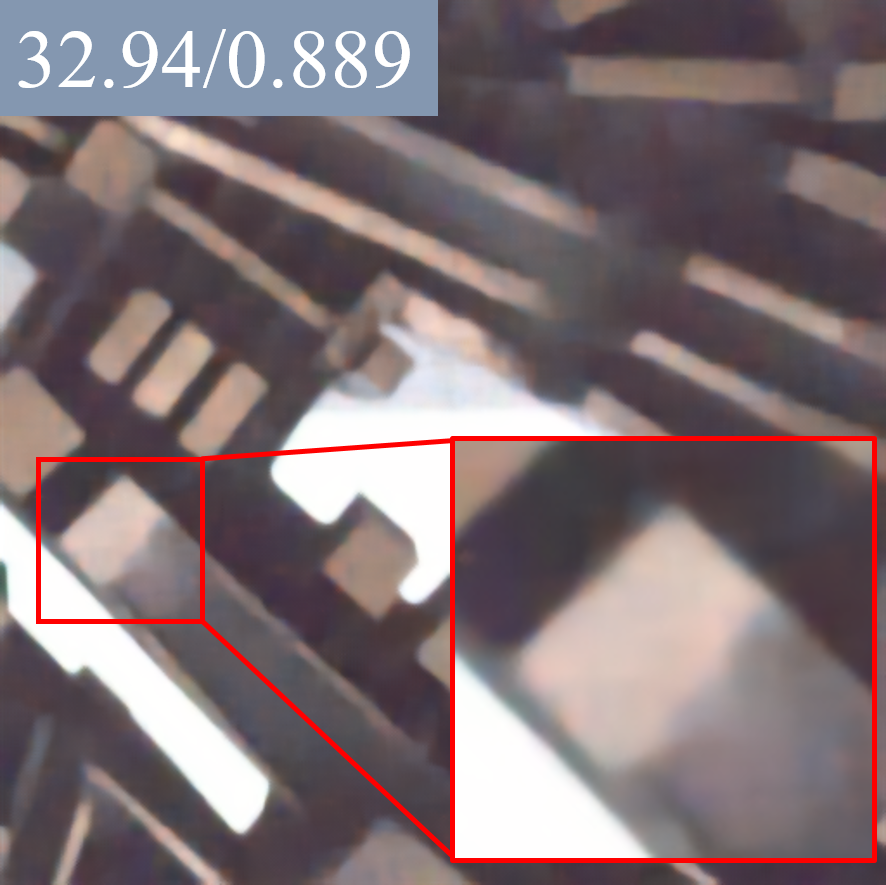}
		\caption{AP-BSN \cite{lee2022ap}\\(Self-supervised)}
		\label{fig:1-c}
	\end{subfigure}
	\hfill
	\begin{subfigure}{0.495\linewidth}
		\includegraphics[width=1\linewidth]{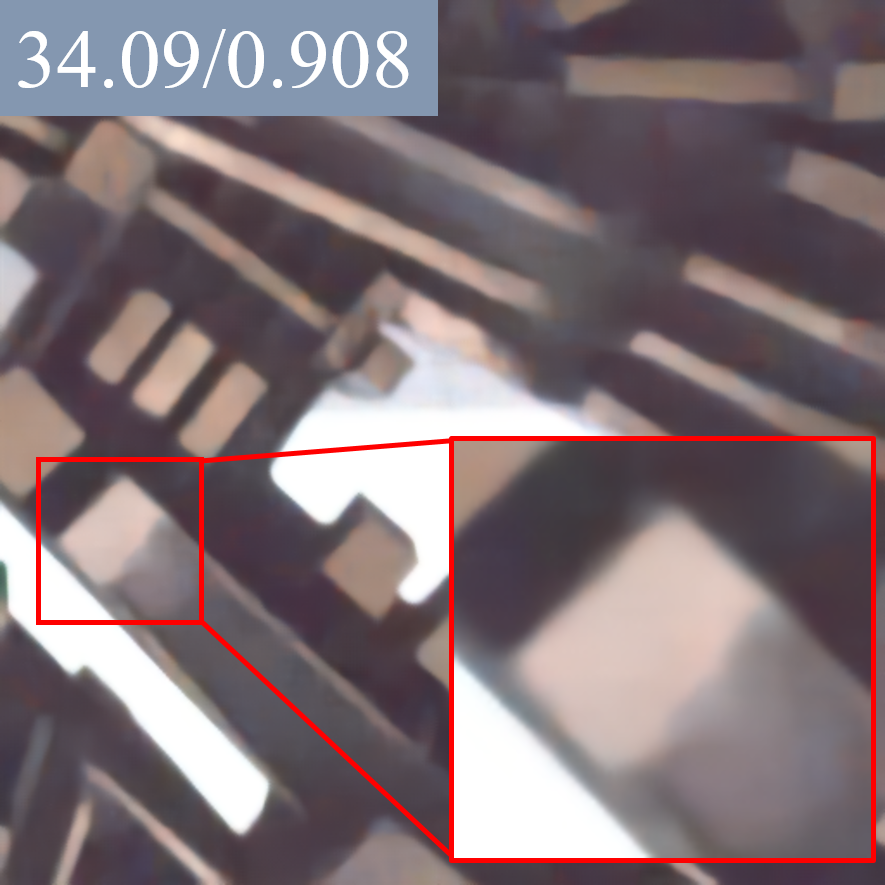}
		\caption{MM-BSN\\(Self-supervised)}
		\label{fig:1-d}
	\end{subfigure}
	\caption{{\bf Visual comparison of our MM-BSN with other competing methods on the DND benchmark.} (a) DnCNN is trained on real-world noisy-clean pairs from the SIDD Medium dataset \cite{abdelhamed2018high}. (b) C2N uses clean SIDD \cite{abdelhamed2018high} and noisy DND \cite{plotz2017benchmarking} samples to simulate the real-world noise distribution in an unsupervised manner. (c) AP-BSN is trained directly on the noisy images in the SIDD Medium dataset \cite{abdelhamed2018high}. (d) MM-BSN is trained on images with real noise from SIDD. We mark the PSNR (dB) and SSIM with respect to the groundtruth for the quantitative comparison.}
	\label{fig:1}
\end{figure}
\setlength{\textfloatsep}{5pt plus 2pt minus 2pt}
\section{Introduction}
Image denoising is a key step in image processing, and the denoising performance has a significant impact on the subsequent image processing tasks. Traditional image denoising methods \cite{dabov2007image,luo2015adaptive,gu2014weighted} are time consuming and costly, but usually have poor robustness in real-world applications. With the advancement of deep learning, learning-based image denoising algorithms have made great progress and can be divided into two classes, supervised methods and self-supervised methods. 

The supervised denoising methods \cite{zhang2017beyond,anwar2019real,chang2020spatial,guo2019toward,yue2019variational,yue2020dual,chen2022simple} have relatively better performance than the self-supervised. However, supervised image denoising requires a large number of noisy-clean image pairs, which are difficult to collect in practical applications, and generating such image pairs requires massive human effort and cost. One of the most common ways is to add simulated real-world noises, such as Additive White Gaussian Noise (AWGN), to clean images to artificially synthesize noisy images so as to obtain synthetic noisy-clean pairs \cite{zhang2017beyond,guo2019toward, zhang2018ffdnet, kim2020transfer, liu2019multi,park2019densely}. Nevertheless, there is always an unavoidable gap between the synthetic noise and the real noise, which severely affects the performance of these supervised models trained on synthetic noise in the real-world image denoising applications. In addition, in some cases, it is also difficult to obtain clean images.

In this situation, many self-supervised image denoising methods \cite{huang2021neighbor2neighbor,zhang2022idr,batson2019noise2self,krull2019noise2void,krull2020probabilistic, lehtinen2018noise2noise} that do not require noisy-clean image pairs have been proposed. Noise2Noise \cite{lehtinen2018noise2noise} used noisy-noisy image pairs to train the model, which achieved comparable performance to supervised algorithms. But it requires two perfectly aligned noisy images, which are difficult to obtain in practice. Noisier2Noise \cite{moran2020noisier2noise} and NAC \cite {xu2020noisy} added the same type of noise as the existing noise to the original noisy image to form noisier-noisy pairs as the training set. This requires the model users to know the specific types of the noise in the image, which is unrealistic in practice because the causes of noise are diverse and the type of noise can change constantly in real-world. IDR \cite{zhang2022idr} adopted an iterative approach, taking the noisy images as inputs to the existing denoising model trained by noisier-noisy pairs, and treating the output as the next round optimization target to further refine the denoising model. In this way, the denoising model is optimized by iterations, which can easily lead to the final denoised image being over-smoothened. Noise2Void \cite{krull2019noise2void} proposed a blind spot network (BSN) denoising method based on the assumption that pixel signals in the image are spatially correlated in the image, and noise signals are spatially independent with zero-mean. In recent years, several publications \cite{krull2019noise2void,laine2019high, honzatko2020efficient,wu2020unpaired} have shown that BSN is effective in synthesizing noise for denoising. However, real-world noise is usually spatially continuous. In most existing BSN denoising methods \cite{batson2019noise2self, krull2019noise2void, krull2020probabilistic, lee2022ap, honzatko2020efficient, wu2020unpaired}, the masks used to generate blind spots have a single pixel blinded in the center, which makes it difficult to denoise when the noise correlated area is large. Zhang et al. \cite{zhang2023self} combined Transformer and CNN to achieve a trade-off between denoising images with global spatially correlated noise and preserving local detail. However, Transformer is computationally intensive, making it difficult to deploy in practical applications on mobile devices \cite{wen2022software}. 

\indent\setlength{\parindent}{1em}Motivated by the fact that different shapes of convolution kernels can extract different features, we propose a variety of masks with different shapes to generate blind spots, such as '+'-shaped mask, '$\square$'-shaped mask, '×'-shaped mask, and so on. The multi-masks with different blind spots are used to mask the surrounding pixels at different positions, so as to destroy the spatial correlations of the noise in multi-direcion. And we systematically demonstrate the effectiveness of using different masks or different mask combinations for image denoising. In addition, we propose an enhanced BSN that combines with the multi-mask strategy, namely MM-BSN, to more efficiently integrate multi-mask paths, recover the destroyed textures, and control the model size. Extensive experiments demonstrate the effectiveness and superiority of the proposed method.

\indent\setlength{\parindent}{1em}The main contributions of our work are as follows:

\indent\setlength{\parindent}{1em}1.To the best of our knowledge, we are the first to explore the combination of different convolution kernels with multi-mask to extract features, and to perform denoising on images with large-scale spatially correlated noise in self-supervised. Furthermore, our multi-mask strategy can be integrated with other methods.

\indent\setlength{\parindent}{1em}2. We propose a novel self-supervised MM-BSN that can integrate the features extracted by multi-masked convolution kernels, control the model size growth, and preserve the image detail when denoising.

\indent\setlength{\parindent}{1em}3. Our approach achieves the state-of-the-art performance among published self-supervised sRGB image denoising methods, which is significant for practical applications.
\section{Related Work}
{\bf Supervised image denoising.} Zhang et al. \cite{zhang2017beyond} first proposed a deep learning based image denoising method called DnCNN, which trained the model with generating noisy-clean pairs by manually adding AWGN to clean images. Subsequently, many  researchers proposed other image denoising methods  \cite{anwar2019real,chang2020spatial, kim2020transfer, fang2020multilevel, lan2021image,liu2019multi,park2019densely, gu2019self} based on deep learning by adding AWGN to clean sRGB images.  However, the denoising performance of these models in the real world was unsatisfactory due to the large gap between artificial and real-world noise. Scholars \cite{brooks2019unprocessing, mildenhall2018burst} proposed to convert sRGB images to rawRGB first, and then added Poisson noise corresponding to shot noise and Gaussian noise corresponding to read noise to rawRGB. After denoising in rawRGB space, the final denoised result image was converted back to sRGB space using ISP tools. For this denoising method, accurate noise estimation and modelling was essential for success. Although the noise obtained by statistical modelling reduced the gap between the synthetic noise and the real noise, the injected noise was not real and external factors could alter the accuracy of the noise modelling. To this end, it was recognized that the most effective way to denoise was to use the noisy-clean pairs \cite{yue2020dual,hu2021pseudo,sharif2020learning,chen2022simple} directly from the real-world when available. However, such a noisy-clean pair dataset requires a huge amount of human labour to collect and a huge amount of time to construct in the real world, and was even more impractical given the diverse application scenarios.

\indent\setlength{\parindent}{1em}{\bf Self-supervised image denoising.} Noise2Noise \cite{lehtinen2018noise2noise} used two perfectly aligned noisy images from the same scene as input and target, respectively. L2 loss was used to minimize the difference between the two noisy images in order to make the model capable of denoising. Then Noise2Void \cite{krull2019noise2void}, Noise2Self \cite{batson2019noise2self}, Probabilistic Noise2Void \cite{krull2020probabilistic}, Neighbor2Neighbor \cite{huang2021neighbor2neighbor}, IDR \cite{zhang2022idr}, CVF-SID \cite{neshatavar2022cvf}, Blind2Unblind \cite{wang2022blind2unblind} and AP-BSN \cite{lee2022ap} were proposed to use only noisy images for training. As the most widely used self-supervised denoising method, BSN was firstly proposed in Noise2Void \cite{krull2020probabilistic}, which is a special CNN that masks pixel in the center of the receptive field, and uses the surrounding information to reconstruct the information of the masked pixels. Its denoising capability is restricted to the assumption that the noise is spatially independent. Noise2Void \cite{krull2020probabilistic} took the masked image as the input and the fully noisy image as the target to train the model. The masked pixels are not used during training, which can easily lead to loss of detail and over-smoothing of the image. Neighbor2Neighbor \cite{huang2021neighbor2neighbor} synthesized two sub-noisy-images by randomly selecting two adjacent pixels from the 4×4 neighbourhood of the rawRGB image. Two sub-noisy-images were used as input and target for training, respectively, forming noisy-noisy pairs. However, training directly on sub-noisy-images would inevitably lose some image detail. To improve this, Blind2Unblind \cite{wang2022blind2unblind} used all the pixels for training by generating sub-masked-images with pixels masked at different positions, and then used a global mask strategy to collect all pixels from the masked positions in the sub-masked-images after denoising. Although Blind2Unblind makes full use of all pixel information, it is difficult to denoise large-noise using only dot-based masks.

Laine19 \cite{laine2019high} occluded half of the receptive fields in four different directions, achieving the effect that the center of the receptive field is not seen. D-BSN \cite{wu2020unpaired} and David et al. \cite{honzatko2020efficient} used the center-masked convolution kernel and the dilated convolution layer (DCL) with a specific step size to construct the BSN. The publications proved that BSN is effective in synthesizing noise for denoising. However, the real-world noise is usually spatially continuous and BSNs would fail to handle it. To break the spatial correlation of real-world noise, AP-BSN  \cite{lee2022ap} adopted Pixel-shuffle Downsampling (PD) with 5-pixel stride on images before training, and utilized center-masked convolution kernel and dilated convolution layer (DCL) to achieve the effect of blind spots during training. However, AP-BSN relies on the PD with limited stride to break the spatial correlation of the noise. If large-noise exists in the image, blindly increasing the PD stride will cause irreversible damage to image details \cite{lee2022ap}. Therefore, it is challenging for AP-BSN to strike a balance between the noise removal and texture information preservation, especially when denoising large-noise. In this paper, we propose a joint feature-extraction method using multi-masked convolutional kernels to destroy large-noise correlations. We also propose a novel architecture that combines the multi-mask convolutional kernels with BSN (MM-BSN) to make full use of the extracted features and preserve texture structures of the original image as much as possible.
\section{Motivation}

\begin{figure}[t]
	\setlength{\abovecaptionskip}{0cm}
	\setlength{\belowcaptionskip}{0cm}
	\centering
	\begin{subfigure}{1\linewidth}
		\includegraphics[width=0.24\linewidth]{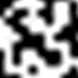}
		\includegraphics[width=0.24\linewidth]{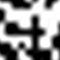}
		\includegraphics[width=0.24\linewidth]{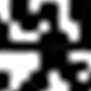}
		\includegraphics[width=0.24\linewidth]{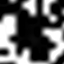}
		\caption{Spatially correlated noise shown by the black area in different shapes.}
		\label{fig:noise1}
	\end{subfigure}
	\begin{subfigure}{0.95\linewidth}
		\includegraphics[width=0.9\linewidth]{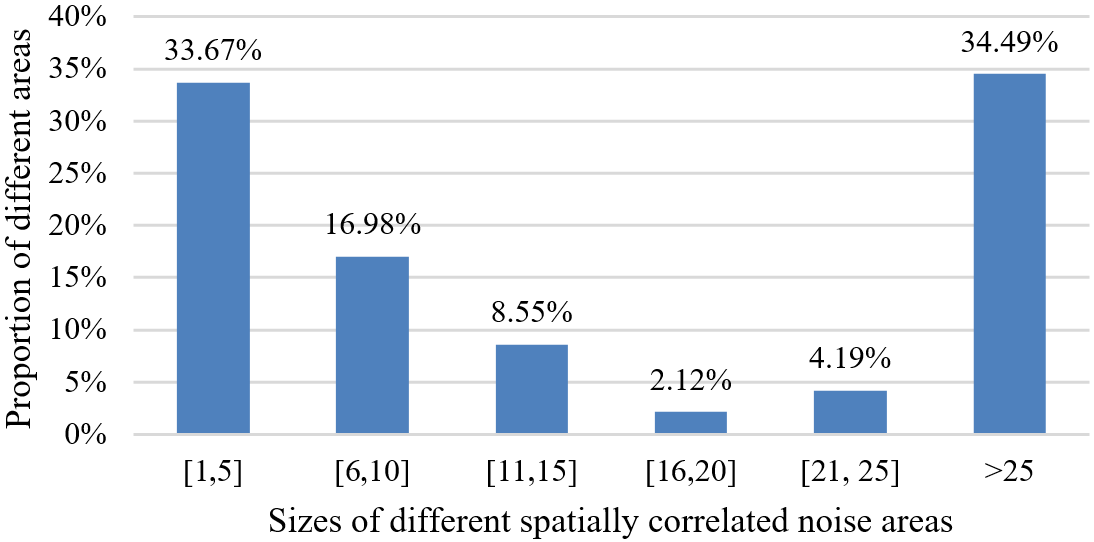}
		\caption{Proportion of different noise area on a full image}
		\label{fig:noise2}
	\end{subfigure}
	\caption{Noise detail on the 0228\underline{ }N.png from SIDD validation.}
	\label{fig:2_1}
\end{figure}

We explore the noise spatially correlations that have different shapes as shown in Figure \ref{fig:2_1}. The sub-images in Figure \ref{fig:noise1} all have a size of height×width as 10×10, which shows that the correlation area is large. We also computes the proportion of spatially correlated noise in different areas of the image in Figure \ref{fig:noise2}. We define the spatially correlated noise with a area bigger than 25 as large-noise. Figure \ref{fig:noise2} shows that the large-noise, which theoretically cannot be handled by PD stride not bigger than 5, occupies more than 1/3. 

Recently published BSN methods, either the mask in the input \cite{huang2021neighbor2neighbor,batson2019noise2self,krull2019noise2void,krull2020probabilistic,wang2022blind2unblind} or the mask in the network \cite{lee2022ap,honzatko2020efficient,wu2020unpaired}, which used a dot-based mask, is not enough to break the correlation of large-noise. In this way, the blind pixels recovered from the surrounding information would still contain noise. Motivated by the prior knowledge that filters with different shapes can be designed to target different types of noise, such as '+', '$\square$', etc., we propose a novel multi-mask strategy, which ultilizes different convolution kernels masked in different shapes to further destroy the spatial connection of noise.

\section{Main Method}

\begin{figure}[t]
	\setlength{\abovecaptionskip}{0cm}
	\setlength{\belowcaptionskip}{0cm}
	\centering
	\begin{subfigure}{0.185\linewidth}
		\includegraphics[width=1\linewidth]{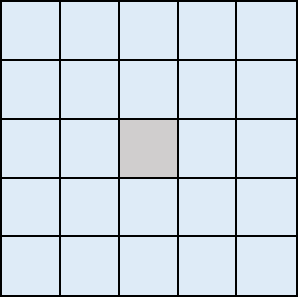}
		\caption{o}
		\label{fig:2-a}
	\end{subfigure}
	\hfill
	\begin{subfigure}{0.185\linewidth}
		\includegraphics[width=1\linewidth]{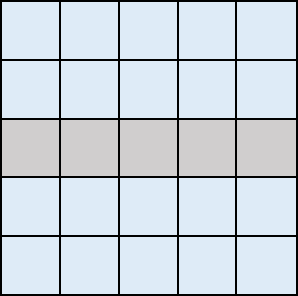}
		\caption{—}
		\label{fig:2-b}
	\end{subfigure}
	\hfill
	\begin{subfigure}{0.185\linewidth}
		\includegraphics[width=1\linewidth]{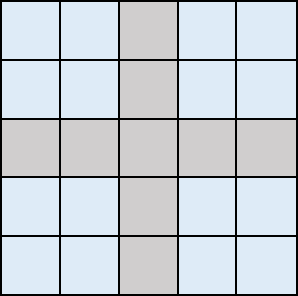}
		\caption{+}
		\label{fig:2-c}
	\end{subfigure}
	\hfill
	\begin{subfigure}{0.185\linewidth}
		\includegraphics[width=1\linewidth]{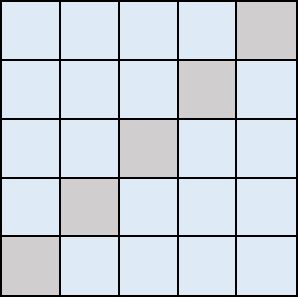}
		\caption{/}
		\label{fig:2-d}
	\end{subfigure}
	\hfill
	\begin{subfigure}{0.185\linewidth}
		\includegraphics[width=1\linewidth]{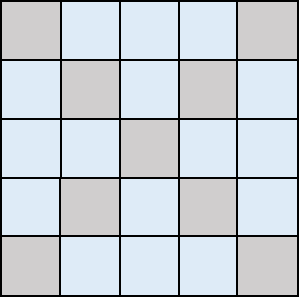}
		\caption{×}
		\label{fig:2-e}
	\end{subfigure}
	
	\begin{subfigure}{0.185\linewidth}
		\includegraphics[width=1\linewidth]{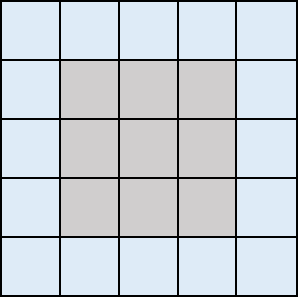}
		\caption{$\square$}
		\label{fig:2-f}
	\end{subfigure}
	\hfill
	\begin{subfigure}{0.185\linewidth}
		\includegraphics[width=1\linewidth]{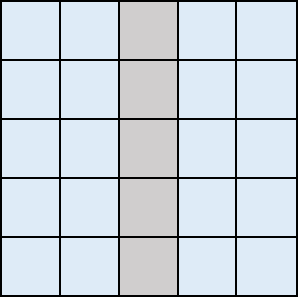}
		\caption{$\vert$}
		\label{fig:2-g}
	\end{subfigure}
	\hfill
	\begin{subfigure}{0.185\linewidth}
		\includegraphics[width=1\linewidth]{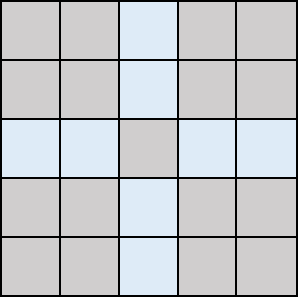}
		\caption{$\boxplus$}
		\label{fig:2-h}
	\end{subfigure}
	\hfill
	\begin{subfigure}{0.185\linewidth}
		\includegraphics[width=1\linewidth]{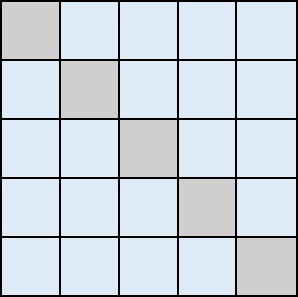}
		\caption{\textbackslash}
		\label{fig:2-i}
	\end{subfigure}
	\hfill
	\begin{subfigure}{0.185\linewidth}
		\includegraphics[width=1\linewidth]{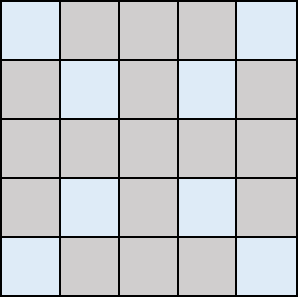}
		\caption{$\divideontimes$}
		\label{fig:2-j}
	\end{subfigure}
	\caption{ {\bf Multi-mask is shown on 5×5 kernel.} Gray dots represent 0, and blue dots represent 1. (a) is the central mask with a single blind spot and (f) is a '$\square$'-shaped mask. (b) is a '—'-shaped mask and (g) is a '$\vert$'-shaped mask. (c) is a '+'-shaped mask and (h) is a '$\boxplus$'-shaped mask. (d) is a '/'-shaped mask and (i) is a '\textbackslash-shaped mask. (e) is a '×'-shaped mask and (j) is a '$\divideontimes$'-shaped mask.}
	\label{fig:2}
\end{figure}

{\bf Multi-Mask Strategy.} We propose to use the multi-mask strategy to further destroy the spatial connection of the noise, while preserving useful texture information of the image.
Figure \ref{fig:2} shows the shapes of different masks when the convolution kernel size is 5×5, such as '+', '$\square$', '—', '$\vert$', '/' , '\verb|\|', '×', etc. 

Theoretically, we can arbitrarily combine multiple masks of different shapes to achieve different denoising results. When using $n$ number of types of masked convolutional kernels types, the same operations are performed for each path until the final concatenation, where all features extracted by several different masked convolutional kernels are fused together. In this way, we can obtain a number of basis multi-mask BSN models, whose architectures contain multiple branches corresponding to the number of masks. However, the model size obtained by this naive method of simple stacking is almost $n$ times the size of the basic network. Consequently, the workload on the hardware device is multiplied by $n$. To control the model size, make full use of the information around the blind spot and avoid information redundancy, we generally use a combination of only two masks. The feature extracted by 'o'-shaped mask contains complete information. However, it may contain more unconducive information for denoising because it could not break the spatial connection of the noise sufficiently. The other types of masks mask more pixels of the surrounding pixels, which can break the spatial connection of the noise more, but lose more image information. So we can combine the feature extracted by 'o' to provide more detail and the other shape of masks to break the spatial correlation of the noise and reinforce each other to get a better denoising performance. Of course, two masks with complementary mask shapes also can break the spatial connection of the noise while extracting information from the surrounding pixels. Multi-mask combinations can be flexibly adjusted according to the real noise distribution. 

Our multi-mask strategy can be integrated with other methods by simply stacking the different mask paths. However, in this way, the increasing number of different mask types will explode the model size. In addition, the features extracted by different masks have no interaction between the processing paths at the intermediate stages before the final concatenation. Without such interaction, information transfer and co-optimization between these processing paths is not possible. Therefore, how to use multi-mask to destroy the spatial connection of noise while retaining more texture information is also a challenge. Last but not the least, as the mask area increases,  the texture information of the image itself is increasingly destroyed. Finally, we propose a novel MM-BSN, to address these challenges.

\begin{figure*}[t]
	\setlength{\abovecaptionskip}{0cm}
	\setlength{\belowcaptionskip}{-10pt}
	\centering
	\includegraphics[width=0.8\linewidth]{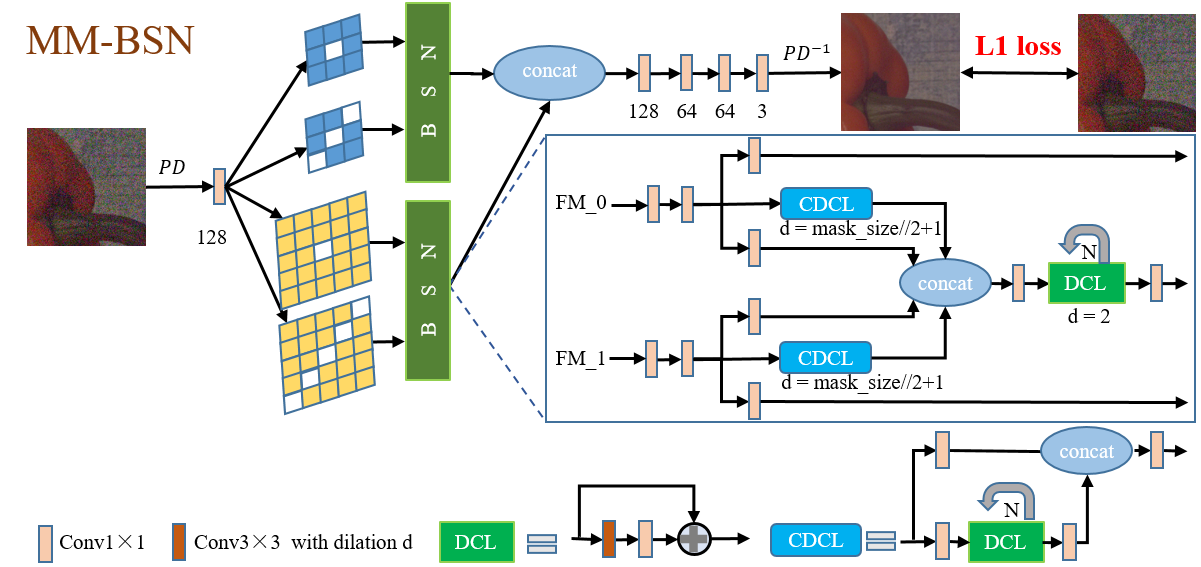}
	\caption{{\bf MM-BSN Architecture.} The channel of feature maps that are not marked is 128.  N indicates that DCL repeats N times.}
	\label{fig:5}
\end{figure*}

\indent\setlength{\parindent}{1em}{\bf MM-BSN Architecture.} MM-BSN is initially motivated by AP-BSN \cite{lee2022ap}. We also use masked convolutional kernels to extract the shallow features. But instead of using only the center mask, we add other shapes of masks to extract the masked features.

\indent\setlength{\parindent}{1em} The architecture of MM-BSN is shown in Figure \ref{fig:5}. The workflow consists of four steps. First, a linear transformation is performed on the noisy image with a 1×1 convolutional layer, and the output feature containing the complete image information passes through several different masked convolutional layers in parallel. Second, each masked feature passes through three layers in parallel, two 1×1 convolutional layers and a Concatenation-based Dilated Convolutional Layer (CDCL). CDCL contains a small number of DCLs (set to 2 in this article) and its output features are combined with one linearly transformed feature  from a 1×1 convolutional layer using a concatenation according to the mask size. The features extracted by the same size but different masked convolution kernels are fused together. Third, after passing through several DCLs (set to 7 in this article), all features are concatenated together, and the features extracted by different masked convolution kernels of different sizes are fused. Finally, the output is obtained by channel transformation and feature fusion with several 1×1 convolutional layers.

Due to the interaction between different feature pathways, the resulting MM-BSN parameter set of 5.3M is larger than AP-BSN of 3.7M, but much smaller than the model size of a simple stack of AP-BSN with multi-mask (namely SMM-BSN) of 7.3M. The ablation experiments of several models are detailed in Section \ref{sec:4.3}.

\indent\setlength{\parindent}{1em}{\bf Loss Chosen.} In this paper, we use L1 loss function to train our MM-BSN:
\begin{equation}
	\setlength{\abovedisplayskip}{0pt}
	\setlength{\belowdisplayskip}{5pt}
	E = \Vert I_{out} - I_{N} \Vert_1
	\label{eq:important}
\end{equation}
\begin{equation}
	\setlength{\belowdisplayskip}{-1em}
	I_{out} = PD^{-1}(M(PD(I_{N})))
	\label{eq:also-important}
\end{equation}
Where $M$ denotes the MM-BSN model, $I_{out}$ is the result of $PD^{-1}$, and $I_{N}$ is the noisy input. Similar to AP-BSN \cite{lee2022ap}, we use PD to preliminarily break the spatial connection between the noises of adjacent pixels. After PD with stride $S_{pd}$, we obtain a group of small sub-images that are inputs to MM-BSN. The denoised result $I_{out}$, which has the same size as the original image, is decoded by operating $PD^{-1}$ to the outputs of the model.
\section{Experiments}
\subsection{Implementation Details}

\setlength{\textfloatsep}{5pt plus 2pt minus 2pt}

\indent\setlength{\parindent}{1em}{\bf Datasets.} We take the public datasets of SIDD \cite{abdelhamed2018high} and DND \cite{plotz2017benchmarking} for our experiments. We take the noisy sRGB images in SIDD Medium dataset that contains 320 pairs of noisy-clean images as the training set and SIDD validation as the valid set, respectively. SIDD validation and SIDD benchmark can be used as test sets. DND dataset that contains only 50 noisy image are generally used as the test dataset. Since only noisy images are needed to train our self-supervised models, we use DND as both the training set and the test set.

{\bf Training Details.} All models are trained with the same hyperparameters. The batch size is 8 and the number of training epochs is 30. The optimization function adopted is Adam. The initial learning rate is 0.0001, and the learning rate of every 8 epochs is multiplied by 0.1. The images are resized to 128×128, and are randomly rotated within a range of 90° in the horizontal or vertical direction before training. All experiments are run on a server with python 3.8.0, pytorch1.12.0, and Nvidia Tesla T4 GPUs. For a relatively fair comparison, unless otherwise stated, we set the PD stride as 5 for training, 2 for testing, and the same post-processing as AP-BSN \cite{lee2022ap}.
\begin{table}[htbp]
	\setlength{\abovecaptionskip}{0pt}
	\setlength{\belowcaptionskip}{-16pt}
	\centering
	\begin{tabular*}{\hsize}{c@{\extracolsep{\fill}}c@{\extracolsep{\fill}}c@{\extracolsep{\fill}}c@{\extracolsep{\fill}}}
		\hline
		& Mask & SIDD Validation	&SIDD Benchmark\\
		\hline
		\multirow{2}*{$S_{pd}$=2}& 'o' & 24.27/0.361	&27.48/0.627  \\
		&'$\square$'	&35.29/0.854	&36.84/0.932\\
		\hline
	\end{tabular*}
	\caption{{\bf Quantitative comparison of the same network using different masks with $S_{pd}$=2.} PSNR/SSIM results are calculated between the denoised-clean pairs using the Python toolkit for SIDD validation, and the official toolkit for SIDD benchmark.}
	\label{tab:table 2}
\end{table}
\begin{figure}[t]
	\setlength{\abovecaptionskip}{0cm}
	\setlength{\belowcaptionskip}{0cm}
	\centering
	\begin{subfigure}{0.24\linewidth}
		\includegraphics[width=1\linewidth]{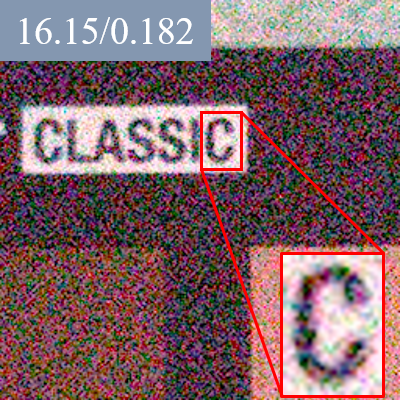}
		\captionsetup{font={scriptsize}}
		\caption{Noisy}
		\label{fig:6-a}
	\end{subfigure}
	\hfill
	\begin{subfigure}{0.24\linewidth}
		\includegraphics[width=1\linewidth]{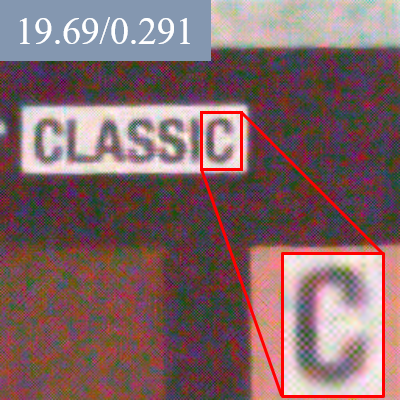}
		\captionsetup{font={scriptsize}}
		\caption{$S_{pd}$=2, 'o'}
		\label{fig:6-b}
	\end{subfigure}
	\begin{subfigure}{0.24\linewidth}
		\includegraphics[width=1\linewidth]{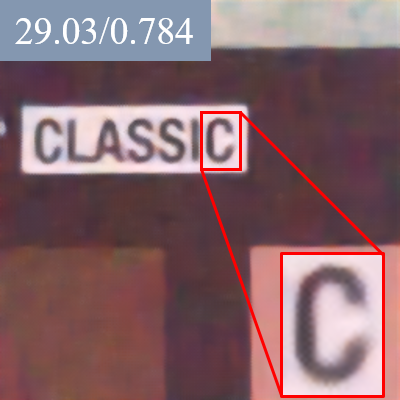}
		\captionsetup{font={scriptsize}}
		\caption{$S_{pd}$=5, 'o'}
		\label{fig:6-c}
	\end{subfigure}
	\hfill
	\begin{subfigure}{0.24\linewidth}
		\includegraphics[width=1\linewidth]{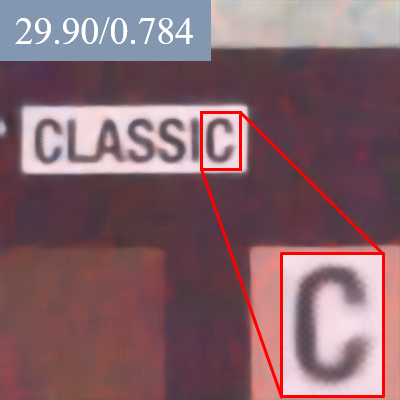}
		\captionsetup{font={scriptsize}}
		\caption{$S_{pd}$=2, '$\square$'}
		\label{fig:6-d}
	\end{subfigure}
	\caption{{\bf Visualization performance of several models with the same architecture but different mask type or different $S_{pd}$.} (a) Noisy image. (b)With a small stride factor $S_{pd}$=2 and center mask, the method cannot remove noise from noisy image.  (c) With  $S_{pd}$=5 and 'o'-shaped mask, the model can denoise better. (d)  With $S_{pd}$=2 and '$\square$'-shaped mask, the model can get a better performance than that with 'o'-shaped mask.}
	\label{fig:6}
\end{figure}
\begin{figure*}[t]
	\setlength{\abovecaptionskip}{0cm}
	\setlength{\belowcaptionskip}{0cm}
	\centering
	\begin{subfigure}{0.18\linewidth}
		\includegraphics[width=3.47cm]{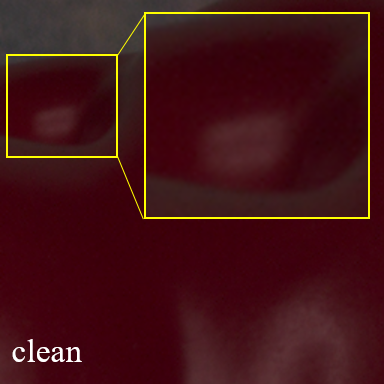}
	\end{subfigure}
	\hfill
	\begin{subfigure}{0.18\linewidth}
		\includegraphics[width=3.47cm]{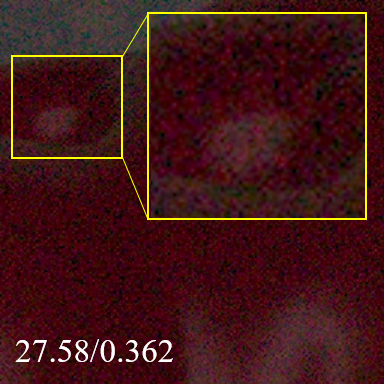}
	\end{subfigure}
	\hfill
	\begin{subfigure}{0.18\linewidth}
		\includegraphics[width=3.47cm]{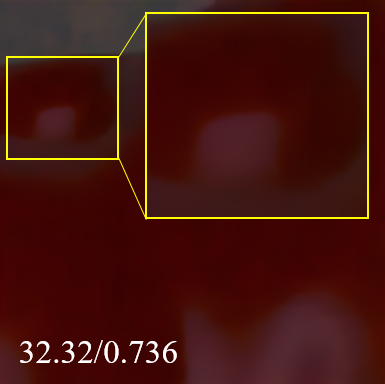}
	\end{subfigure}
	\hfill
	\begin{subfigure}{0.18\linewidth}
		\includegraphics[width=3.47cm]{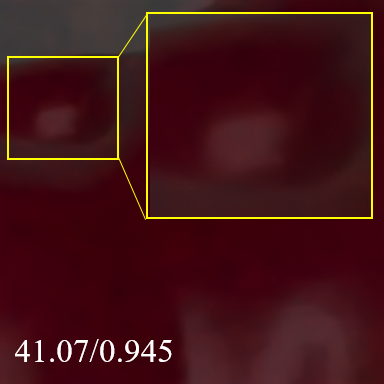}
	\end{subfigure}
	\hfill
	\begin{subfigure}{0.18\linewidth}
		\includegraphics[width=3.47cm]{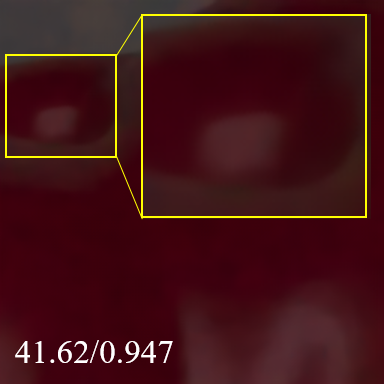}
	\end{subfigure}
	\begin{subfigure}{0.18\linewidth}
		\includegraphics[width=3.47cm]{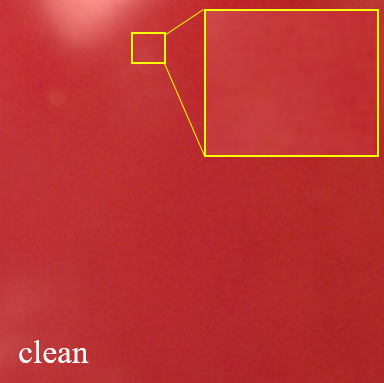}
		\caption{Clean}
		\label{fig:4-a2}
	\end{subfigure}
	\hfill
	\begin{subfigure}{0.18\linewidth}
		\includegraphics[width=3.47cm]{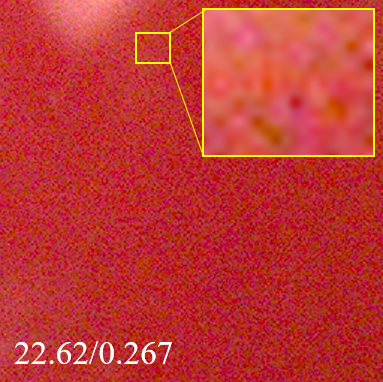}
		\caption{Noisy}
		\label{fig:4-b2}
	\end{subfigure}
	\hfill
	\begin{subfigure}{0.18\linewidth}
		\includegraphics[width=3.47cm]{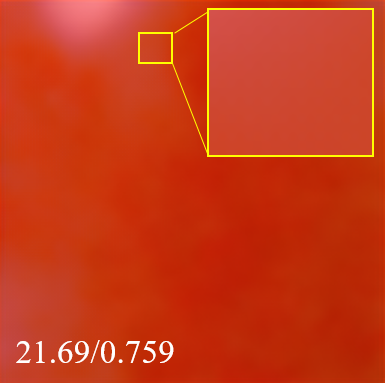}
		\caption{AP-BSN}
		\label{fig:4-c2}
	\end{subfigure}
	\hfill
	\begin{subfigure}{0.18\linewidth}
		\includegraphics[width=3.47cm]{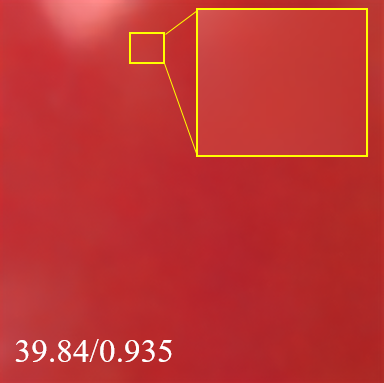}
		\caption{'o' + '+'}
		\label{fig:4-d2}
	\end{subfigure}
	\hfill
	\begin{subfigure}{0.18\linewidth}
		\includegraphics[width=3.47cm]{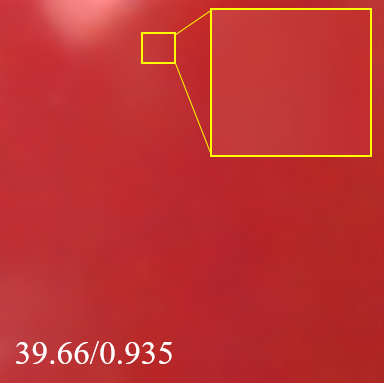}
		\caption{'o' + '$\boxplus$' + '+'}
		\label{fig:4-e2}
	\end{subfigure}
	\setlength{\belowcaptionskip}{-1em}
	\caption{{\bf Qualitative comparison of several methods with different mask combinations on SIDD validation dataset.} (a) Clean images. (b) Noisy images. (c) Denoised images by AP-BSN \cite{lee2022ap}. (d) Denoised images by SMM-BSN using a combination of 'o' mask and '+' mask. (e) Denoised images by SMM-BSN using a combination of 'o', '$\boxplus$' and '+'-shaped mask.} 
	\label{fig:4}
\end{figure*}
\subsection{Analyzing Multi-Mask strategy in BSN}

To compare the performance of the proposed method with different mask combinations, we trained several MM-BSN models with different masks on SIDD Medium dataset. All trained models are quantitatively evaluated on SIDD validation and benchmark. Existing Python toolkits are used to compute the PSNR/SSIM of SIDD validation. At the same time, we upload the denoised results of SIDD benchmark  to the official website and obtain the reported PSNR/SSIM.

{\bf Significant effect on breaking the noise structure.} To check the effectiveness of the mask in breaking the structure of large-noise, we take the image after PD with $S_{pd}$=2 as input to train.  Figure \ref{fig:6} shows the denoising performance of the models with different $S_{pd}$s or masks but the same other settings. AP-BSN \cite{lee2022ap} shows that when $S_{pd}$=2, the spatial connection of the noise in the image cannot be broken well. Using only the center mask, the model is weakly able to denoise, as shown in Figure \ref{fig:6-b}. But if we use the '$\square$' mask when $S_{pd}$=2, the denoised result is even better than the 'o'-shaped masked model with $S_{pd}$=5 \cite{lee2022ap} as shown in Figure \ref{fig:6-c} and \ref{fig:6-d}. Table \ref{tab:table 2} quantitatively shows that, the PSNR/SSIM on the SIDD validation and benchmark datasets are greatly improved when the center mask is replaced with the '$\square$' mask when $S_{pd}$=2. This proves that the large-noise structure can be better broken by the '$\square$' mask, and it is not only an exception shown in  Figure \ref{fig:6}, but also a common case.

{\bf Quantitative comparison of MM-BSN models.} We summarize the following points from Table \ref{tab:table 1}: (1) Different mask combinations achieve different denoising performance. The reason is that different masks target different noise correlations, and it is common sense that the final denoised result will be different. (2) The mask combinations combined with 'o' are overall better than the combinations without it, due to that the feature extracted by the 'o'-shaped mask preserves the texture information of the image itself more completely. (3) The combination of '/' and '\verb|\|' gives the best performance, followed by the combination of 'o' and '/', which indicates that the dataset has more '/' and '\verb|\|' shaped spatially related noise. For different datasets, the noise structures are different, and users can freely choose the combination of masks or design the mask shape suitable for the real dataset according to the needs.

\begin{table}[t]
	\setlength{\abovecaptionskip}{0cm}
	\centering
	\begin{tabular*}{\hsize}{c@{\extracolsep{\fill}}c@{\extracolsep{\fill}}c@{\extracolsep{\fill}}c@{\extracolsep{\fill}}c@{\extracolsep{\fill}}c@{\extracolsep{\fill}}c@{\extracolsep{\fill}}c@{\extracolsep{\fill}}c@{\extracolsep{\fill}}l@{\extracolsep{\fill}}l@{\extracolsep{\fill}}}
		\hline
		\multicolumn{9}{c}{Masks} & \multicolumn{2}{c}{Test datasets} \\
		\hline
		o & — & $\vert$ & $\boxplus$ & + &/ & \verb|\| & $\divideontimes$ & × & Validation & Benchmark\\
		\checkmark & \checkmark &  &  &  &  &  &  &  &                  37.34/\underline{0.881}  &    37.31/{\bf 0.937}\\
		\checkmark &	 & \checkmark &  &  &  &  &  &  & 	     37.30/\underline{0.881} &    37.30/{\bf 0.937}\\
		\checkmark & &  &\checkmark  &  &  &  &  &  &         37.32/{\bf 0.882}	&  37.31/\underline{0.936}\\
		\checkmark	& & &	 &\checkmark  &  &  &  &  &		  37.28/0.879 &	37.28/\underline{0.936}\\
		\checkmark	& & & &	& \checkmark &  &  &  &		  \underline{37.37}/{\bf 0.882} &	\underline{37.35/0.936}\\
		\checkmark	& & & & & & \checkmark &  &  &		      37.18/0.878	&37.18/0.935\\
		\checkmark	& & & & & &	& \checkmark &  &			  37.24/\underline{0.881}	&37.23/0.934\\
		\checkmark	& & & & & & & &\checkmark  &  		      37.24/\underline{0.881}	&37.24/0.934\\
		& \checkmark &\checkmark& & & & & &	      &37.12/0.879  &37.12/0.934 \\
		& & & \checkmark & \checkmark&	& & & &		          37.19/0.880	&37.18/0.934\\
		& & & & &\checkmark	&\checkmark	& & & 		          {\bf 37.38/0.882}   &{\bf 37.37}/\underline{0.936}\\
		& & & & & & &\checkmark	&\checkmark &		      37.11/0.879	&37.11/0.933\\
		\hline
	\end{tabular*}
	\setlength{\belowcaptionskip}{-8pt}
	\caption{{\bf Quantitative comparison of MM-BSN with different mask combinations on SIDD validation and benchmark datasets with PSNR/SSIM.}}
	\label{tab:table 1}
\end{table}

\indent\setlength{\parindent}{1em}{\bf Comparison of BSN models with increasing mask types.} Figure \ref{fig:4} shows the qualitative denoising performance of models with different number of mask types in the same framework on the SIDD validation dataset. Comparing Figure \ref{fig:4-c2}, \ref{fig:4-d2} and \ref{fig:4-e2}, it can be observed that by adding other types of masks based on the 'o'-shaped mask, the denoising performance is significantly improved. Especially in the second row, the denoised result of AP-BSN has unacceptable color shifts, while SMM-BSN restores the original color perfectly, indicating that adding masks of other shapes can effectively destroy the noise correlation during feature extraction. In addition, it can be seen from the second row of Figure \ref{fig:4-b2} and Figure \ref{fig:4-c2} that the PSNR value after denoising decreases from 22.62dB to 21.69dB when only the center mask is used, but it increases to 39.84dB/39.66dB when the multi-mask is used. This indicates that when the spatially correlated  noise region is large, the center mask alone cannot break the noise structure sufficiently. Since the features extracted by the center mask alone may still be noisy, the final denoising result will be biased by massive noise. Figure \ref{fig:4-d2} and \ref{fig:4-e2} show that increasing the number of mask types does not always improve the denoising performance. The possible reason for this is that features extracted by increasing types of masks lead to the information redundancy, which is unsensive and unuseful for denoising. 

\begin{table}[t]
	\setlength{\abovecaptionskip}{0cm}
	\setlength{\belowcaptionskip}{-5pt}
	\centering
	\begin{tabular*}{\hsize}{c@{\extracolsep{\fill}}c@{\extracolsep{\fill}}c@{}}
		\toprule
		Models & SIDD Validation & Parameters(M)\\
		\midrule
		AP-BSN & 35.91/0.870 & 3.7 \\
		SMM-BSN &	37.16/0.879 & 7.3 \\
		MM-BSN	& {\bf 37.38/0.882} &	5.3 \\
		\bottomrule
	\end{tabular*}
	\caption{{\bf Comparison of Several BSNs.}  All BSNs are trained on SIDD Medium dataset. PSNR/SSIM results for SIDD validation and the model size are shown here.}
	\label{tab:table 4}
\end{table}

\begin{figure}[t]
	\setlength{\abovecaptionskip}{0cm}
	\setlength{\belowcaptionskip}{0cm}
	\centering
	\begin{subfigure}{0.32\linewidth}
		\includegraphics[width=1\linewidth]{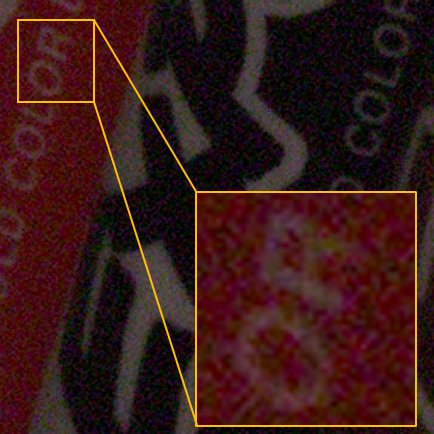}
		\caption{Noisy}
		\label{fig:7-a}
	\end{subfigure}
	\begin{subfigure}{0.32\linewidth}
		\includegraphics[width=1\linewidth]{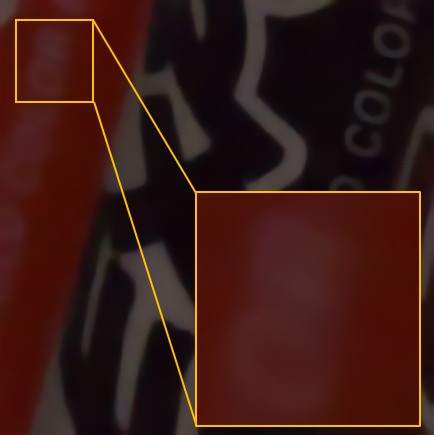}
		\caption{AP-BSN \cite{lee2022ap}}
		\label{fig:7-b}
	\end{subfigure}
	\begin{subfigure}{0.32\linewidth}
		\includegraphics[width=1\linewidth]{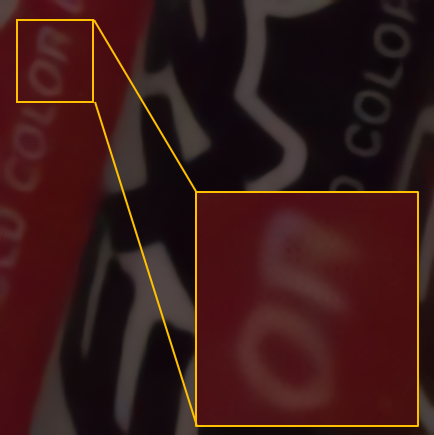}
		\caption{MM-BSN}
		\label{fig:7-c}
	\end{subfigure}
	\setlength{\belowcaptionskip}{-5pt}
	\caption{{\bf Visual comparison of AP-BSN and MM-BSN on the SIDD benchmark.} They are trained on SIDD Medium dataset using the same center mask. (a) Noisy image. (b) Denoised result by AP-BSN \cite{lee2022ap}. (c) Denoised result by MM-BSN.}
	\label{fig:7}
\end{figure}

\begin{figure}[t]
	\setlength{\abovecaptionskip}{0cm}
	\setlength{\belowcaptionskip}{0cm}
	\centering
	\begin{subfigure}{1\linewidth}
		\includegraphics[width=1\linewidth]{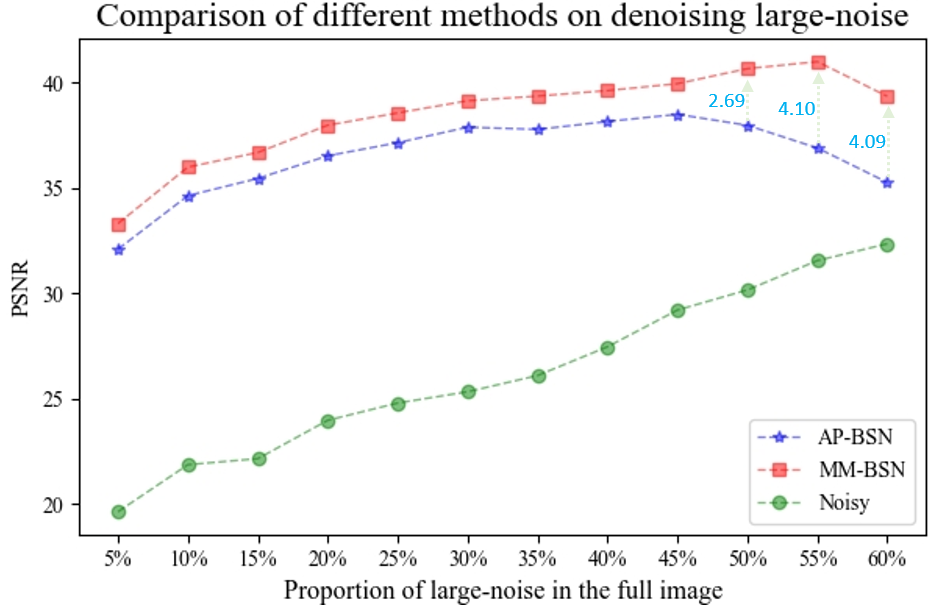}
	\end{subfigure}
	\caption{{\bf Comparison of denoising results of AP-BSN and MM-BSN on the noisy sets of SIDD Validation.}}
	\label{fig:9}
\end{figure}

\begin{table*}
	\setlength{\abovecaptionskip}{0pt}
	\setlength{\belowcaptionskip}{0pt}
	\centering
	\begin{tabular*}{\hsize}{c@{\extracolsep{\fill}}l@{\extracolsep{\fill}}l@{\extracolsep{\fill}}l@{\extracolsep{\fill}}l@{\extracolsep{\fill}}l@{\extracolsep{\fill}}}
		\hline
		\multirow{2}*{ } & \multirow{2}*{Method} & \multicolumn{2}{c}{SIDD} & \multicolumn{2}{c}{DND} \\
		&   & PSNR & SSIM &PSNR  &SSIM \\
		\hline
		\multirow{2}*{Non-learning based}&BM3D \cite{dabov2007image} & 25.65 & 0.685	& 34.51 & 0.851 \\
		& WNNM \cite{gu2014weighted} & {\bf 25.78} &	{\bf 0.809}	&{\bf 34.67}&	{\bf 0.865}\\
		\hline
		\multirow{2}*{\makecell[c]{Supervised\\Synthetic pairs}}&DnCNN \cite{zhang2017beyond} & 23.66&	0.583&	32.43&	0.790 \\
		&CBDNet \cite{guo2019toward} &	{\bf 33.28}&	{\bf 0.868}	&{\bf 38.05}	&{\bf 0.942}\\
		\hline
		\multirow{4}*{\makecell[c]{Supervised\\Real pairs}}&DnCNN \cite{zhang2017beyond} 	& 36.07\textsuperscript{$\diamond$} &	0.911\textsuperscript{$\diamond$} & 37.81\textsuperscript{$\diamond$} &	0.931\textsuperscript{$\diamond$}\\
		&DnCNN \cite{zhang2017beyond} &	36.07&	0.911& 37.81& 	0.931\\
		&AINDNet(R)\textsuperscript{$\ast$}\cite{kim2020transfer} 	&38.84	&0.951	&39.34	&0.952\\
		&VDN \cite{yue2019variational} 	&39.26	&0.955	&39.38	&0.952\\
		&NAFNet \cite{chen2022simple} 	&{\bf 40.30}	&{\bf 0.962}	&-	&-\\
		\hline
		\multirow{3}*{\makecell[c]{Unsupervised\\Unpaired}} &GCBD \cite{chen2018image} &	-	&-	&35.58	&0.922\\
		&C2N \cite{jang2021c2n} + DIDN\textsuperscript{$\ast$} \cite{yu2019deep} 	&{\bf 35.35}	&{\bf 0.937}	&37.28	&0.924\\
		&D-BSN \cite{wu2020unpaired} + MWCNN \cite{liu2018multi} 	&- 	&- 	&{\bf 37.93}	&{\bf 0.937}\\
		\hline
		\multirow{12}*{\makecell[c]{Self-supervised}}& Noise2Void \cite{krull2019noise2void} &	27.68\textsuperscript{\bf R}	&0.668\textsuperscript{\bf R}	&-	&-\\
		&Noise2Self \cite{batson2019noise2self} 	&29.56\textsuperscript{\bf R}	&0.808\textsuperscript{\bf R}	&-	&-\\
		&NAC \cite{xu2020noisy} 	          &-	&-	&36.20	&0.925\\
		&R2R \cite{pang2021recorrupted}   	&34.78	&0.898	&-	&-\\
		&CVF-SID (S2) \cite{neshatavar2022cvf}	&34.71	&0.917	&36.50	&0.924\\
		&AP-BSN \cite{lee2022ap}	&35.97	&0.925	&38.09	&0.937\\
		&AP-BSN\textsuperscript{$\dagger$} \cite{lee2022ap}	&36.91	&0.931	&-	&-\\
		&MM-BSN(Ours)	     &{\bf 37.37}	& {\bf 0.936}	&{\bf 38.46}	&{\bf 0.940}\\
		&MM-BSN\textsuperscript{$\dagger$}(Ours)	&-	&-	& {\bf 38.74}	& {\bf 0.943}\\
		\hline
	\end{tabular*}
	\caption{{\bf Quantitative comparison of different denoising models on SIDD and DND benchmarks.} By default, we get the official evaluation results from SIDD and DND benchmark websites. $\diamond$ indicates that we have retrained the model, uploaded the test results and received the results. {\bf R} indicates that the result is reported by R2R \cite{pang2021recorrupted}. $\ast$ denotes the method with self-ensemble strategy \cite{lim2017enhanced}. $\dagger$ denotes the model trained with the same training and test data sets. The highest value is highlighted in {\bf bold} for each type of denoising model.}
	\label{tab:table 3}
\end{table*}

\subsection{Analyzing our network architecture}
For fairly comparing, all models are trained on SIDD Medium dataset and evaluated on SIDD validation. AP-BSN \cite{lee2022ap} uses only the center mask, SMM-BSN and MM-BSN use the combination of '/'-shaped mask and '\verb|\|'-shaped mask for training, and other settings are the same as before.

\begin{figure*}
	\setlength{\abovecaptionskip}{0cm}
	\setlength{\belowcaptionskip}{0cm}
	\centering
	\begin{subfigure}{0.18\linewidth}
		\includegraphics[width=3.47cm]{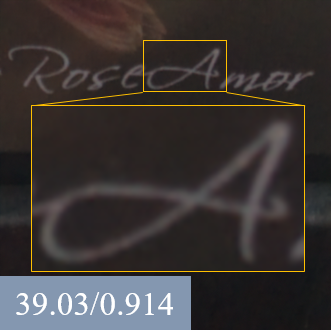}
	\end{subfigure}
	\hfill
	\begin{subfigure}{0.18\linewidth}
		\includegraphics[width=3.47cm]{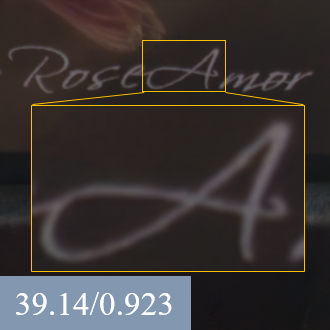}
	\end{subfigure}
	\hfill
	\begin{subfigure}{0.18\linewidth}
		\includegraphics[width=3.47cm]{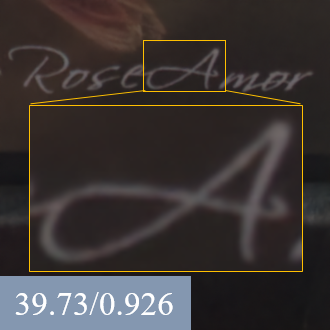}
	\end{subfigure}
	\hfill
	\begin{subfigure}{0.18\linewidth}
		\includegraphics[width=3.47cm]{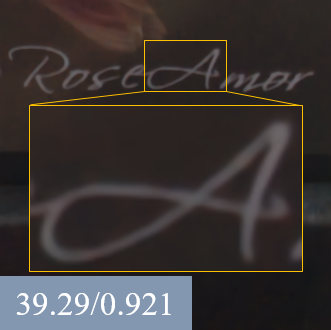}
	\end{subfigure}
	\hfill
	\begin{subfigure}{0.18\linewidth}
		\includegraphics[width=3.47cm]{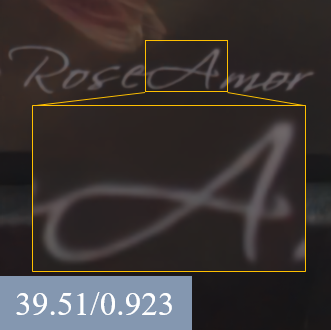}
	\end{subfigure}
	\begin{subfigure}{0.18\linewidth}
		\includegraphics[width=3.47cm]{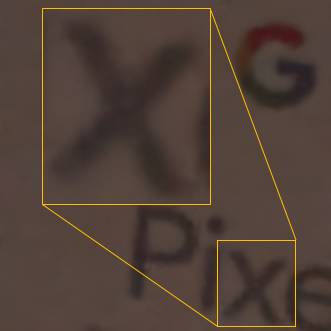}
		\captionsetup{justification=centering}
		\caption{DnCNN \cite{zhang2017beyond} \\Supervised-Real SIDD}
		\label{fig:8-b2}
	\end{subfigure}
	\hfill
	\begin{subfigure}{0.18\linewidth}
		\includegraphics[width=3.47cm]{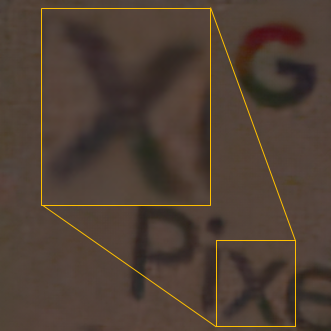}
		\captionsetup{justification=centering}
		\caption{C2N\cite{jang2021c2n}+DIDN\textsuperscript{$\ast$}\cite{yu2019deep}\\ Unpaired}
		\label{fig:8-c2}
	\end{subfigure}
	\hfill
	\begin{subfigure}{0.18\linewidth}
		\includegraphics[width=3.47cm]{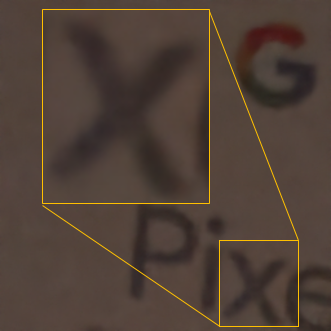}
		\captionsetup{justification=centering}
		\caption{CBDNet \cite{guo2019toward}\\Supervised-Synthetic noise}
		\label{fig:8-d2}
	\end{subfigure}
	\hfill
	\begin{subfigure}{0.18\linewidth}
		\includegraphics[width=3.47cm]{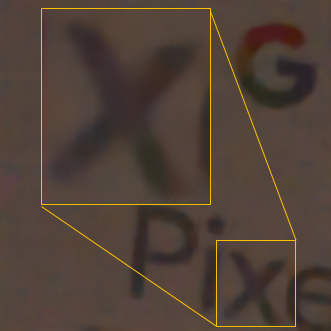}
		\captionsetup{justification=centering}
		\caption{AP-BSN \cite{lee2022ap}\\Self-supervised}
		\label{fig:8-e2}
	\end{subfigure}
	\hfill
	\begin{subfigure}{0.18\linewidth}
		\includegraphics[width=3.47cm]{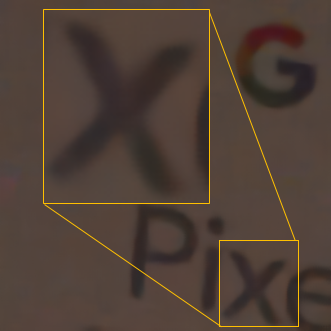}
		\captionsetup{justification=centering}
		\caption{MM-BSN\\Self-supervised}
		\label{fig:8-f2}
	\end{subfigure}
	\setlength{\belowcaptionskip}{-10pt}
	\caption{{\bf Qualitative comparison between different denoising methods on SIDD and DND benchmarks.} (a) DnCNN is trained on the real paired SIDD Medium dataset. (b) C2N generates a realistic noisy image from the clean input, where the following denoising model, i.e., DIDN, is trained on the generated pairs. (c) CBDNet is trained in a supervised manner using noisy-clean pairs, where the noisy image is obtained by adding synthetic noise to the clean image. (d-e) The methods are trained directly on real sRGB images. Note that the DND benchmark (upper) provides some per-sample PSNR/SSIMs, while SIDD benchmark (lower) does not.}
	
	\label{fig:8}
\end{figure*}
Table \ref{tab:table 4} compares AP-BSN and its corresponding extended versions SMM-BSN, which indicates that the denoising performance is significantly improved by applying the multi-mask strategy. The PSNR/SSIM of SIDD validation shows that MM-BSN (37.38/0.882) outperforms AP-BSN (35.91/0.870) by a large margin. Figure \ref{fig:7-b} shows that using AP-BSN, the alphabets in the yellow box of the image are blurred and a lot of detail is lost. Figure \ref{fig:7-c} shows that the alphabets in the yellow box are generally preserved. This observation suggests that by adding concatenation-based skip-connections from the shallow features in MM-BSN, the lost detail can be supplemented in time. 

We classify images with different large-noise ratios for SIDD validation and calculate the average PSNR of AP-BSN and MM-BSN in each image set, as shown in Figure \ref{fig:9}. Our MM-BSN with multi-mask strategy outperforms AP-BSN with only 'o' masks by a large magin, with PSNR improvements of up to 4, particularly in large-noise.
\label{sec:4.3}

\subsection{MM-BSN in real-world sRGB image denosing}
The proposed MM-BSN aims to denoise large-noise in sRGB images by combining multi-mask in the self-supervised manner, while preserving the texture detail and controlling the model size.

Table \ref{tab:table 3} quantitatively compares the denoising performance of several traditional algorithms, supervised denoising algorithms, unsupervised and self-supervised algorithms on SSID and DND benchmarks. The table shows that MM-BSN performs best in self-supervised methods and even outperforms some supervised algorithms. Furthermore, our MM-BSN does not require rawRGB images and noise estimation like R2R, nor real noisy-clean pairs like supervised models. Therefore, in practical applications, researchers can  train MM-BSN directly on the noisy images from the target scene for denoising, avoiding degradation of the model performance when the scenario changes. 

Figure \ref{fig:8} qualitatively compares the visual denoising performance of state-of-the-art models on a random image in SIDD and DND benchmarks. Compared with its yellow box in the upper images, the lines area denoised by self-supervised models in Figure \ref{fig:8-e2} and Figure \ref{fig:8-f2} are more smoothing, while there are unwanted but obvious burring effects near the lines denoised by other models shown in Figure \ref{fig:8-b2}, Figure \ref{fig:8-c2} and Figure \ref{fig:8-d2}. MM-BSN performs better than most of the models, and can even compete with the supervised method of CBDNet \cite{guo2019toward} with slightly lower PSNR/SSIM. Comparing the yellow box in the lower images, our MM-BSN has a more clear boundary contour of outer boundary of the alphabets, and the noise on the alphabets themselves is more obviously reduced.  

\section{Conclusion}
In this paper, we propose a multi-mask strategy worked on BSNs for self-supervised sRGB image denoising. Multi-mask can significantly break the large-noise structure, which previously cannot be efficiently handled by the only-center-masked models. In addition, we develop MM-BSN to effectively combine the features extracted by multi-masked convolutional layers and control the model size to grow without explosion. In particular, the utilization of concatenation-based skip-connections can help to compensate for the loss of information caused by the masks. Extensive experiments prove that our method can effectively denoise the images with a large scale spatially correlated noise and can preserve more textures, achieving a better denoising performance than other unsupervised and self-supervised methods in the literature. Our proposed MM-BSN is well suited for a real practical application scenario considering that it only needs noisy sRGB images to train.

{\small
	\bibliographystyle{ieee_fullname}
	\bibliography{my_ref}
}
\end{document}